\pdfoutput=1

\documentclass[11pt]{article}
\usepackage[final]{acl}

\usepackage{times}
\usepackage{latexsym}
\usepackage{multirow}
\usepackage{enumitem}

\usepackage[T1]{fontenc}

\usepackage[utf8]{inputenc}

\usepackage{microtype}

\usepackage{inconsolata}

\usepackage{graphicx}

\usepackage{algpseudocode}
\usepackage{algorithm}
\usepackage{amsmath}
\usepackage{booktabs}
\usepackage{amssymb}
\usepackage{titlesec}
\titlespacing{\section}{0pt}{1.5ex plus 0.5ex minus .2ex}{0.8ex}
\titlespacing{\subsection}{0pt}{1.25ex plus 0.5ex minus .2ex}{0.6ex}

\title{DAVIS: Planning Agent with Knowledge Graph-Powered Inner Monologue}

\author{
 \textbf{Minh Pham-Dinh\textsuperscript{1}},
 \textbf{Munira Syed\textsuperscript{2}},
 \textbf{Michael Yankoski\textsuperscript{1,3}},
 \textbf{Trenton W. Ford\textsuperscript{3}}
\\
\\
 \textsuperscript{1}Davis Institute for Artificial Intelligence, Colby College, United States \\
 \textsuperscript{2}University of Notre Dame, United States \\
 \textsuperscript{3}William \& Mary, United States \\
 \\
 \small{
   \textbf{Correspondence:} \href{mailto:mhpham26@colby.edu}{mhpham26@colby.edu}
 }
}

\begin{document}
\maketitle

\begin{abstract}

Designing a generalist scientific agent capable of performing tasks in laboratory settings to assist researchers has become a key goal in recent AI research. Unlike everyday tasks, scientific tasks are inherently more delicate and complex, requiring agents to possess a higher level of reasoning ability, structured and temporal understanding of their environment, and a strong emphasis on safety. Existing approaches often fail to address these multifaceted requirements. To tackle these challenges, we present DAVIS\footnote{All code and prompts are available on \href{https://github.com/minhphd/DAVIS}{Github}}. Unlike traditional retrieval-augmented generation (RAG) approaches, DAVIS incorporates structured and temporal memory, which enables model-based planning. Additionally, DAVIS implements an agentic, multi-turn retrieval system, similar to a human's inner monologue, allowing for a greater degree of reasoning over past experiences. DAVIS demonstrates substantially improved performance on the ScienceWorld benchmark compared to previous approaches on 8 out of 9 elementary science subjects. In addition, DAVIS's World Model demonstrates competitive performance on the famous HotpotQA and Musique dataset for multi-hop question answering. To the best of our knowledge, DAVIS is the first RAG agent to employ an interactive retrieval method in a RAG pipeline.

\end{abstract}
\begin{figure}
    \centering
    \includegraphics[width=0.80\linewidth]{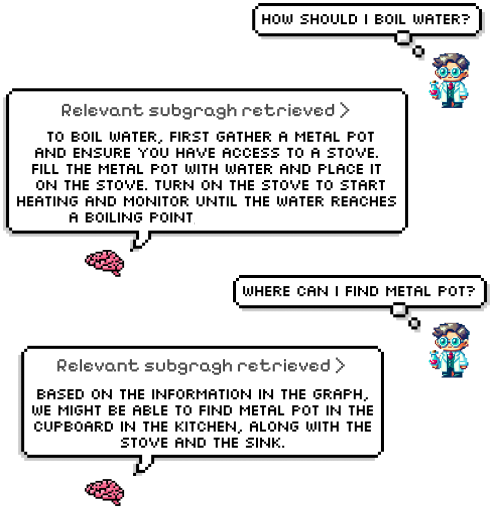}
    \caption{Visualization of DAVIS's inner monologue during decision-making. This process involves iteratively querying its knowledge graph to fill knowledge gaps in order to plan its next action.}
    \label{fig:qa}
\end{figure}

\begin{figure*}
    \centering
    \includegraphics[width=0.80\linewidth]{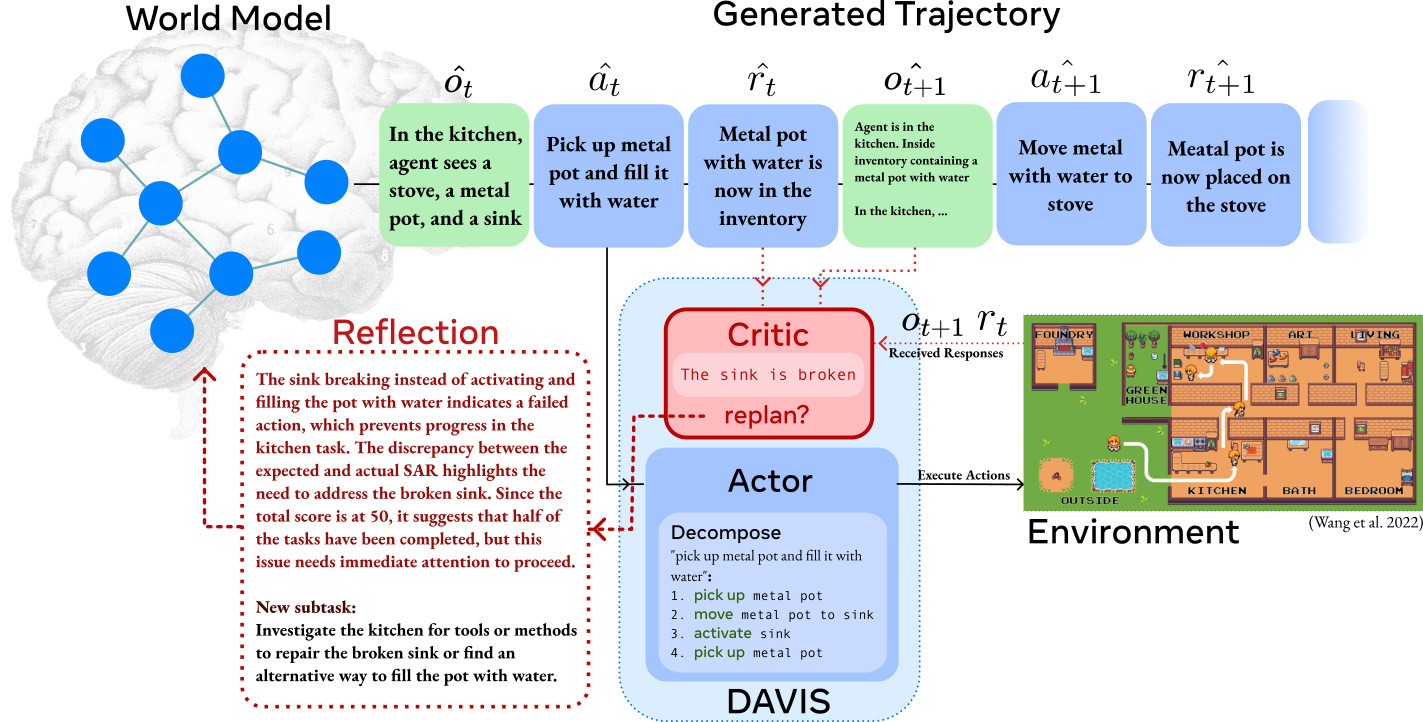}
  \caption[Caption for LOF]{Overview of DAVIS’s decision-making process. The World Model generates a feasible course of actions, which are translated by the actor and executed sequentially by the agent in the environment. The critic detects discrepancies between expected and actual outcomes, identify failures, and suggest replanning. \footnotemark}

    \label{fig:main-dia}
\end{figure*}

\section{Introduction}
A core focus of current Artificial Intelligence (AI) research is the development of artificial agents capable of autonomously performing human tasks requiring high decision-making autonomy \cite{ahn_as_2022, zhao_empowering_2024, wang_describe_2024, putta_agent_2024}. While Reinforcement Learning (RL) has traditionally been used to create goal-oriented agents in Markovian environments \cite{mnih_playing_2013, schrittwieser_mastering_2020, hafner_dream_2020}, it often suffers from sample inefficiency, limited generalizability, and poor interpretability, making real-world deployment challenging \cite{dulac-arnold_challenges_2019}. Recently, large language models (LLMs) \cite{radford_improving_nodate, touvron_llama_2023} have revolutionized the creation of autonomous agents by leveraging natural language understanding to enhance interpretability and generalization. These LLM-based agents have shown great promise in critical domains such as healthcare (\citealp{qiu_llm-based_2024}) and scientific research (\citealp{schmidgall_agent_2025}) by mimicking human decision-making processes and enabling more intuitive reasoning and actions.  \footnotetext{Original ScienceWorld graphic incorporated into this figure under \href{https://creativecommons.org/licenses/by-sa/4.0/}{CC BY-SA 4.0}.}

Several approaches have enhanced agentic reasoning and decision-making. SwiftSage \cite{lin_swiftsage_2023} emulates the fast and slow thinking of humans with fine-tuned language models for planning. SayCan \cite{ahn_as_2022} decomposes tasks into subgoals, while ReAct \cite{yao_react_2023} \footnote{SwiftSage, Reflexion, SayCan, and ReAct are used under MIT license} integrates reasoning into execution. RAG-based systems like Reflexion \cite{shinn_reflexion_2023} and RAP\footnote{RAP is used under MIT license} \cite{kagaya_rap_2024} retrieve past experiences via semantic search, but their unstructured memory limits multi-hop reasoning and causal understanding. These systems retrieve static information rather than engaging in agentic, multi-turn retrieval, preventing dynamic adaptation.

Humans do not retrieve past knowledge statically; instead, we actively reflect, question our understanding, and refine our knowledge through internal dialogues. Inspired by this, we introduce DAVIS, an agentic multi-turn retrieval system that mirrors human cognition by enabling iterative interactions between the agent and its memory during the planning stage, a process we call \textit{inner monologue}. DAVIS actively engages with its World Model (WM), a temporal knowledge graph-based QA system, to refine its understanding before planning and execution. DAVIS engages in conversation with its WM to retrieve past experiences, evaluate actions, identify gaps, and optimize strategies.


DAVIS proves to be effective for iterative reasoning within scientific domains. Specifically, DAVIS outperforms 4 other baselines \cite{ahn_as_2022, kagaya_rap_2024, yao_react_2023, shinn_reflexion_2023} on 8 out of 9 science subjects in the ScienceWorld \cite{wang_scienceworld_2022} environment.\footnote{ScienceWorld is used under Apache 2.0 license} DAVIS's WM achieves competitive performance on the HotpotQA \cite{yang_hotpotqa_2018} and MusiqueQA \cite{trivedi_musique_2022} dataset \footnote{The HotpotQA dataset is distributed under the CC BY-SA 4.0 license. The MusiqueQA dataset is distributed under CC BY 4.0 license}. Our contributions can be summarized as follows:

\begin{itemize}[left=0pt, itemsep=4pt, parsep=0pt, topsep=4pt]
    \item We introduce \textbf{DAVIS}, an agentic reasoning framework that leverages multi-turn retrieval and self-reflection to improve decision-making.
    \item Unlike static retrieval methods, DAVIS leverages a structured temporal knowledge graph memory system to enable multi-hop reasoning and causal understanding.
    \item Empirical evaluations show that DAVIS outperforms prior agentic reasoning models across scientific benchmarks, demonstrating superior planning and execution.
\end{itemize}

\begin{figure*}[t]
    \centering
    \includegraphics[width=0.85\linewidth]{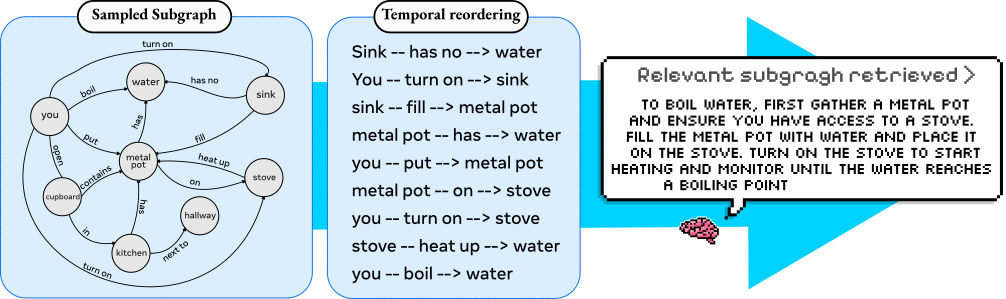}
    \caption{DAVIS’s retrieval and reasoning process. Left: subgraph with relevant entities and their relationships. Middle: temporal reordering of the retrieved information to establish a coherent sequence of actions. Right: DAVIS generates a structured and interpretable response.}
    \label{fig:retrieval-process}
\end{figure*}
\section{Background \& Related Work}
\label{sec:background}
\subsection{Text-Based Scientific Environments}

Text-based simulation environments provide a flexible, language-driven interface for training intelligent agents in open-ended settings. Notably, TextWorld \cite{cote_textworld_2019} offers structured tasks within interactive fiction environments, enabling agents to learn through inventory tracking, natural language feedback, and symbolic state transitions. Its lightweight interface supports high-throughput simulation, with TextWorldExpress achieving up to 4 million steps per second \cite{jansen_textworldexpress_2023}, making it an ideal testbed for large-scale training.

Building on this paradigm, ScienceWorld extends text-based simulation to procedural scientific tasks. Unlike typical interactive fiction environments, ScienceWorld emphasizes tasks that mirror real-world laboratory scenarios, such as growing plants, operating microscopes, or conducting chemical experiments. These tasks introduce a set of multifaceted requirements:

\begin{enumerate}
    \item \label{req:reasoning} \textbf{Multi-hop scientific reasoning} over domain-specific procedural knowledge;
    \item \label{req:temporal} A \textbf{temporal and structured understanding} of how the environment evolves over time, especially when actions have delayed or compounding effects;
    \item \label{req:safety} An emphasis on \textbf{safety and interpretability}, where agents must reason about physical consequences and justify their behavior;
    \item \label{req:observability} Operation under \textbf{partial observability}, requiring internal modeling to compensate for limited real-time feedback.
\end{enumerate}

DAVIS is designed to explicitly address these challenges. In Section~\ref{sec:DAVIS}, we show how each architectural module supports one or more of these multifaceted requirements.

\subsection{LLM agentic systems}
Recent advancements in LLM-based agentic systems have drawn heavily from human decision-making processes and generally fall into two paradigms: direct interaction via chain-of-thought (CoT) reasoning or Retrieval-Augmented Generation (RAG).

The first paradigm involves agents interacting directly with their environment using CoT reasoning \cite{yao_react_2023, ahn_as_2022, lin_swiftsage_2023}. Chain-of-Thought prompting (\citealp{wei_chain--thought_2023}) enables large language models to decompose complex tasks into smaller, interpretable reasoning steps. However, CoT-based systems lack robust memory for long-term learning and adaptability across multiple tasks. The absence of memory has been linked to increased hallucination and stochasticity in task planning (\citealp{guerreiro_hallucinations_2023}), posing risks in domains such as scientific research.

The second paradigm, RAG-based systems, integrates retrieval mechanisms with generative capabilities, enabling agents to access relevant external knowledge during task execution. In the Minecraft domain, extensive work has been done on RAG-based agents, with JARVIS-1 \cite{wang_jarvis-1_2023} and Voyager \cite{wang_voyager_2023} representing the state-of-the-art. Since Minecraft is one of the most popular video games in the world, these agents leverage the extensive in-domain knowledge of LLMs but face significant limitations in scientific environments, where tasks often involve unknown skills and cannot rely on pre-existing knowledge. In such cases, a more general and iterative approach involving multiple trials is necessary.

Recent systems such as Reflexion \cite{shinn_reflexion_2023} and RAP \cite{kagaya_rap_2024} address some of CoT’s limitations by incorporating episodic memory and semantic retrieval. Reflexion leverages trial histories to guide decision-making, while RAP uses nearest-neighbor search to ground decisions in past experience. However, both approaches rely on unstructured vector databases, which fragment information and limit the agent’s ability to perform multi-hop or causal reasoning. Furthermore, they lack temporal modeling capabilities and do not support iterative plan refinement. While prior frameworks such as \textit{Reflexion} use the term “actor” to denote a module that generates an entire action trajectory upfront, followed by a separate “evaluator” or “critic” that assesses success post hoc, DAVIS departs from this sequential model. Instead, DAVIS adopts an actor-critic architecture that operates \textit{during} task execution. The actor grounds high-level plans into fine-grained actions in the environment, while the critic continuously evaluates each step by comparing real-time observations to expectations derived from the plan and temporal knowledge graph. In light of these limitations, there is a growing need for hybrid systems that integrate structured memory, iterative retrieval, and real-time validation. DAVIS addresses this gap by combining a temporal knowledge graph with multi-turn agentic reasoning and critic-guided control, offering a more interpretable and adaptable framework for complex scientific environments.

\subsection{Graph Question Answering (Graph QA)}
Graph Question Answering (Graph QA) systems have become effective tools for structured reasoning and information retrieval. GraphReader \cite{li_graphreader_2024} constructs a graph from document chunks and deploys an agent for exploration. HOLMES \cite{panda_holmes_2024} extracts relevant documents, builds an entity-document graph, prunes it, and uses cosine similarity for answers. GraphRAG \cite{edge_local_2024} generates an entity knowledge graph, pregenerates community summaries, and synthesizes responses. By encoding knowledge in a graph format, these systems excel at multi-hop reasoning over interconnected concepts, making them particularly valuable for domains that require relational understanding, such as scientific research. Unlike unstructured vector-based retrieval systems, Graph QA systems enable iterative retrieval, allowing agents to retrieve information, reason over it, and perform subsequent queries based on the refined context.

\section{DAVIS}
\label{sec:DAVIS}
DAVIS adopts a model-based planning approach \cite{sutton_reinforcement_1998}, where the agent uses a World Model (WM) as an internal representation of its surrounding environment. 
\subsection{Problem Formulation}
\label{sec:formulation}
We define the planning problem for DAVIS in a textual environment as a Partially Observable Markov Decision Process (POMDP): 

\[
\mathcal{P} = (\mathcal{S}, \mathcal{A}, \mathcal{T}, \mathcal{R}, \Omega, \mathcal{O}, \gamma)
\]

In this formulation, \( \mathcal{S} \) denotes the set of true environment states, which are not directly observable. \( \mathcal{A} \) represents the set of available actions. \( \mathcal{T}(s_{t+1} \mid s_t, a_t) \) is the state transition probability function, modeling the dynamics of the environment. \( \mathcal{R}(s_t, a_t) \) is the reward function, specifying the immediate reward received after taking action \( a_t \) in state \( s_t \). \( \Omega \) is the set of possible observations. \( \mathcal{O}(o_{t+1} \mid s_{t+1}, a_t) \) is the observation probability function, defining the likelihood of observing \( o_{t+1} \) given the new state \( s_{t+1} \) and action \( a_t \). \( \gamma \in [0,1) \) is the discount factor, determining the present value of future rewards.

Since the true state \( s_t \) is not directly observable, the agent maintains a belief state \( b_t \), which is a probability distribution over all possible states, representing the agent's estimate of the environment's state at time \( t \). The belief state is updated based on the agent's actions and received observations. The agent selects an action \( a_t \in \mathcal{A} \) based on its current belief state, following a policy \( \pi \):
\[
a_t = \pi(b_t)
\]
After executing the action \( a_t \), the agent receives a reward \( r_t = \mathcal{R}(s_t, a_t) \) and transitions to a new state \( s_{t+1} \) according to the transition function \( \mathcal{T} \). The objective of the agent is to find an optimal policy \( \pi^* \) that maximizes the expected cumulative discounted reward over time:
\[
\pi^* = \arg\max_\pi \mathbb{E} \left[ \sum_{t=0}^\infty \gamma^t r_t \mid \pi \right]
\]

\begin{algorithm}[t]
\small
\caption{Planning with Retrieval-Augmented World Model}
\label{planning}
\textbf{Input}: $\tau$, $\mathcal{R}$ \quad  
\textbf{Parameters}: $L$, $k$ \quad  
\textbf{Output}: $\tau$ \\
\vspace{-0.15in}
\begin{algorithmic}[1]
\For{$t = 1$ to $L$}
    \State $\hat{s}_t \gets f(\tau)$ \quad \Comment{State estimation}
    \State $\hat{a}_t \gets \pi(\hat{s}_t, \mathcal{R}, k)$
    \State $\tau \gets \tau \cup \hat{a}_t$
    \State $\hat{o}_{t+1}, \hat{r}_{t+1} \gets \Call{Transition}{\hat{s}_t, \hat{a}_t}$ \Comment{Algorithm~\ref{transition_predict}}
    \State $\tau \gets \tau \cup \{\hat{o}_{t+1}, \hat{r}_{t+1}\}$
    \If{$(\tau)$ violates safety constraints (optional)} 
        \State Alert supervisor
    \EndIf
\EndFor
\State \Return $\tau$
\end{algorithmic}
\end{algorithm}

\subsection{World Model (WM)}
The World Model (WM) of DAVIS is represented as a Temporal Knowledge Graph (TKG), constructed through a combination of Stanford CoreNLP\footnote{We used default hyperparameters provided by the Stanza package for CoreNLP} \cite{manning_stanford_2014} for coreference resolution and LLM prompting for knowledge extraction. In textual environments, where state representations are conveyed in natural language, constructing an effective WM requires methods that can process and represent textual information efficiently and accurately.

State representation methods in text-based environments include text encoding techniques using recurrent neural networks (\citealp{narasimhan_language_2015}, \citealp{he_deep_2016}, \citealp{hausknecht_interactive_2020}), transformers \cite{kim_plm-based_2022}, and knowledge graph (KG) representations \cite{ammanabrolu_graph_2020}. KGs offer structured and interpretable representations without requiring extensive training. \citeposs{ammanabrolu_learning_2021} framed KG construction in text-based games as a question-answering problem, where agents identified objects and their attributes. This approach demonstrated that higher-quality KGs led to improved control policies. DAVIS generalizes this paradigm by introducing Temporal KGs, which explicitly incorporate time-sensitive information to model environment dynamics. Rather than capturing a static snapshot of world knowledge, the TKG continuously evolves with the agent’s interactions. This enables causal and temporal reasoning: agents can infer the effects of specific actions, reason over state transitions, and plan with respect to how entities evolve over time, addressing the temporal and structured understanding requirement. 

Temporal reasoning is critical in such settings and, as noted in \cite{lee_temporal_2023}, LLMs are highly effective in extrapolating TKGs using in-context learning.

Let \( G_t = (\mathcal{E}, \mathcal{R}, \mathrm{T}) \) denote the Temporal Knowledge Graph (TKG) at time \( t \), where \(\mathcal{E}\) is the set of entities at $t$, \(\mathcal{R}\) is the set of relations representing relationships between entities at $t$, and \(\mathrm{T}\) is the set of timestamps associated with each relation \( e_i \).

During training, when DAVIS executes an action \( a_t \) and receives the subsequent observation \( o_{t+1} \), the transition is stored as:
\[
(o_t \mathbin\Vert a_t \mathbin\Vert o_{t+1})
\]
We prompted an LLM to summarize the concatenated transition and applied Stanford CoreNLP for coreference resolution. The resolved text is then analyzed to extract entities \( V_t \) and relations tuples using LLM-based parsing.

Each extracted tuple \( (v_i, e_j, v_k, \tau) \) is added to the TKG, where the timestamp \( \tau \) records the time at which the fact was introduced:
\[
G_{t+1} = G_t \cup \{(v_i, e_j, v_k, \tau)\}
\]

\subsection{Retrieval-Augmented Model Approximation}
As demonstrated in \citeposs{lee_temporal_2023}, LLMs excel at recognizing temporal patterns and extrapolating future events based on past data. DAVIS leverages this capability to approximate future states and rewards. For example, if sufficient past data indicates that opening a cupboard often reveals a kettle, the LLM can infer such transitions purely from learned patterns without requiring explicit pre-programmed rules. Unlike prior works (\citealp{kagaya_rap_2024, shinn_reflexion_2023}) that rely on vector-based retrieval of experiences, DAVIS employs a more agentic approach. DAVIS engages in an internal conversational process with its graph-powered WM, a process we term \textit{inner monologue}. This process involves iteratively querying its knowledge graph to fill knowledge gaps while retrieving relevant subgraphs to generate informed responses. The graph-powered inner monologue retrieval system is described in Section~\ref{subsec:agentic_retrieval}. 

Although the true state \( s_t \) is not directly observable as mentioned in Section~\ref{sec:formulation}, it is theoretically possible to maintain a statistic $f(\tau)$ that approximates the belief state from the trajectory history. The statistic is updated recurrently and captures all relevant information necessary for optimal decision-making \cite{nguyen_belief-grounded_2021, astrom_optimal_1965}. Applying this to DAVIS, we approximate the belief state \(\hat{b}_t\) with equation:
\[
\hat{b}_t = f(\tau_{t':t}),
\]
where \( f(\cdot) \) is a prompted LLM that extracts relevant information from the trajectory history and is updated recurrently with new observations and actions. To further refine decision-making, DAVIS maintains an inner monologue \( \mathcal{M}_t \), a running list of iterative queries and answers exchanged between DAVIS and its WM, as illustrated in Figure~\ref{fig:qa}. This monologue allows the system to dynamically update its WM based on retrieved insights.

DAVIS optimizes its policy while simultaneously learning approximations of the transition and reward models using its WM. The learned functions incorporating the inner monologue are:
\begin{align}
    \text{Policy:} & \quad \pi(a_t \mid \hat{b}_t, \mathcal{M}_t) \label{eq:action_model} \\
    \text{Transition Model:} & \quad \hat{\mathcal{T}}(o_{t+1} \mid \hat{b}_t, a_t, \mathcal{M}_t) \label{eq:transition_model} \\
    \text{Reward Model:} & \quad \hat{\mathcal{R}}(r_t \mid \hat{b}_t, a_t, \mathcal{M}_t) \label{eq:reward_model}
\end{align}

With the approximated belief \( \hat{b}_t \), DAVIS's WM estimates the transition and reward models using prior experiences retrieved from a TKG. DAVIS leverages previous experiences to directly inform its policy as defined in Equation~\eqref{eq:action_model}. This retrieval-driven approximation enables DAVIS to construct an adaptive and context-aware model of the world, allowing for informed decision-making in complex, temporally dependent environments through multi-hop reasoning.

\subsection{Inner Monologue Retrieval System}
\label{subsec:agentic_retrieval}
Given a query \( q \), such as \textit{"Where can I find water?"}, the WM first narrows its search to relevant entity types such as \texttt{Person (PER)} and \texttt{Location (LOC).} It then selects the two most relevant entities from the available options. Limiting the scope to two entities is computationally efficient and ensures a manageable search space without sacrificing relevant context. The query is then expanded and processed as follows and illustrated in Figure~\ref{fig:retrieval-process}:

\begin{enumerate}[left=0pt, itemsep=4pt, parsep=0pt, topsep=4pt]
    \item \textbf{We iteratively expand} the current list of selected entities by adding their neighbors, forming a maximal subgraph, as ignoring temporal information might result in an infeasible path. 
    \item \textbf{We reorder the edges} in the maximal subgraph based on timestamps. This reordering shows the proper sequence of events. 
    \item \textbf{The temporal sequence is then passed to an LLM} as in-context examples for extrapolation and summarization, enabling the LLM to generate a coherent response. 
\end{enumerate}

\subsection{Planning and Execution with a WM}
With the reward model and transition model approximated, we can now plan action trajectories within the WM. Algorithm~\ref{planning} describes the WM-incorporated planning process used in DAVIS.

\begin{algorithm}[t]
\small
\caption{Transition Prediction}
\label{transition_predict}
\textbf{Input}: $\hat{b}_t$, $\hat{a}_t$, $k$ \quad  
\textbf{Output}: $\hat{o}_{t+1}$, $\hat{r}_{t+1}$ \\
\vspace{-0.15in}
\begin{algorithmic}[1]
\State $\mathcal{M} \gets \emptyset$ \quad \Comment{Initialize inner monologue set}
\State $i \gets 0$
\While{$i < k$ or not \texttt{predicted}}
    \State $\hat{o}_{t+1}, q \gets \hat{T}(\hat{b}_t, \hat{a}_t, \mathcal{M})$
    \State $\hat{r}_{t+1}, q \gets \hat{R}(\hat{b}_t, \hat{a}_t, \mathcal{M})$
    \If{$q \neq \emptyset$}
        \State $\mathcal{M} \gets \mathcal{M} \cup \{(q, \texttt{graphQA}(q))\}$
    \EndIf
    \State $i \gets i + 1$
\EndWhile
\State \Return $\hat{o}_{t+1}, \hat{r}_{t+1}$
\end{algorithmic}
\end{algorithm}

For plan execution, we employ an actor-critic structure, consisting of two distinct models: the actor \( R_a \) and the critic \( R_c \), both integrated with the WM architecture. The process is illustrated in Figure~\ref{fig:main-dia}.

\paragraph{World Model (WM). } The primary objective of the WM is to generate a comprehensive plan or trajectory for achieving a specific task within the environment. Given an initial observation estimate \( \hat{o}_t \), the WM generates a predicted trajectory 
\[\tau_{t:t+L} = \big\{ (\hat{o}_i, \hat{a}_i, \hat{o}_{i+1},  \hat{r}_{i+1} ) \big\}_{i=t}^{t+L-1}
\]
of length \( L \). This trajectory \( \tau_{t:t+L} \) is passed to the actor-critic model for execution in the environment. The WM generates high-level, natural language plans that outline the intended sequence of actions. These plans are explicit, interpretable, and can be validated prior to execution, allowing the agent to reason in advance about safety and correctness. Thus, this module addresses the safety and interpretability requirement.

\paragraph{Actor. } The actor \( R_a \) decomposes each high-level action \( \hat{a}_t \in \tau \) into executable commands within the given environment domain. It also predicts intermediate state transitions between actions:
\[
\hat{\tau}_{t:t+L'} = R_a(\tau_{t:t+L})
\]
where \( L' \geq L \) accounts for the expanded trajectory with executable low-level actions. The actor model is prompted with permissible commands in the current environment. After decomposition, the expanded trajectory \( \hat{\tau}_{t:t+L'} \) is executed step-by-step in the environment, producing actual environment responses:
\[
(o_t, r_t, o_{t+1}) = \mathcal{E}(\hat{a}_t)
\]
where \( \mathcal{E} \) is the environment transition function that maps the executed action \( \hat{a}_t \) to the resulting observation \( o_{t+1} \) and reward \( r_t \). These results are passed to the critic model.


\paragraph{Critic. } The critic \( R_c \) evaluates the actual execution results against the predicted trajectory \( \tau \). The comparison is performed through an LLM-based evaluation function, which assesses the semantic consistency between the expected and actual observations. At each timestep \( t \), the critic receives the predicted state transition \( (\hat{o}_t, \hat{r}_t, \hat{o}_{t+1}) \) and the actual environment response \( (o_t, r_t, o_{t+1}) \) obtained from executing \( \hat{a}_t \) in the environment.

The LLM-based critic compares these components via a prompted evaluation function \(R_c\):
\[
\Delta_t = R_c \Big( (\hat{o}_t, \hat{r}_t, \hat{o}_{t+1}), (o_t, r_t, o_{t+1}) \Big)
\]
where \( \Delta_t \) is a qualitative feedback score representing the level of agreement between the predicted and actual transitions. We called this reflection as the agent "reflects" on the differences between predictions and actual environment responses. Based on the LLM's response, the critic determines whether replanning is necessary. If the predicted and actual observations deviate significantly, the critic updates its list of reflections  \( \mathfrak{R}_t \) (which the world model uses as additional context for future replannings within the current execution) and then triggers replanning.
\[
\mathfrak{R}_{t+1} = \mathfrak{R}_t \cup \{(o_t, \hat{s}_t, \Delta_t)\}
\]
Algorithm~\ref{planning} is then called to replan the new subtask. For example, if the task is "using the stove to heat water" and the agent encounters an exception (e.g., the stove is broken), the LLM evaluates the exception, updates \( \mathcal{M}_t \), and suggests a revised subtask such as "find an alternative heating method."

\begin{figure*}[t]
    \centering
    \includegraphics[width=0.85\linewidth]{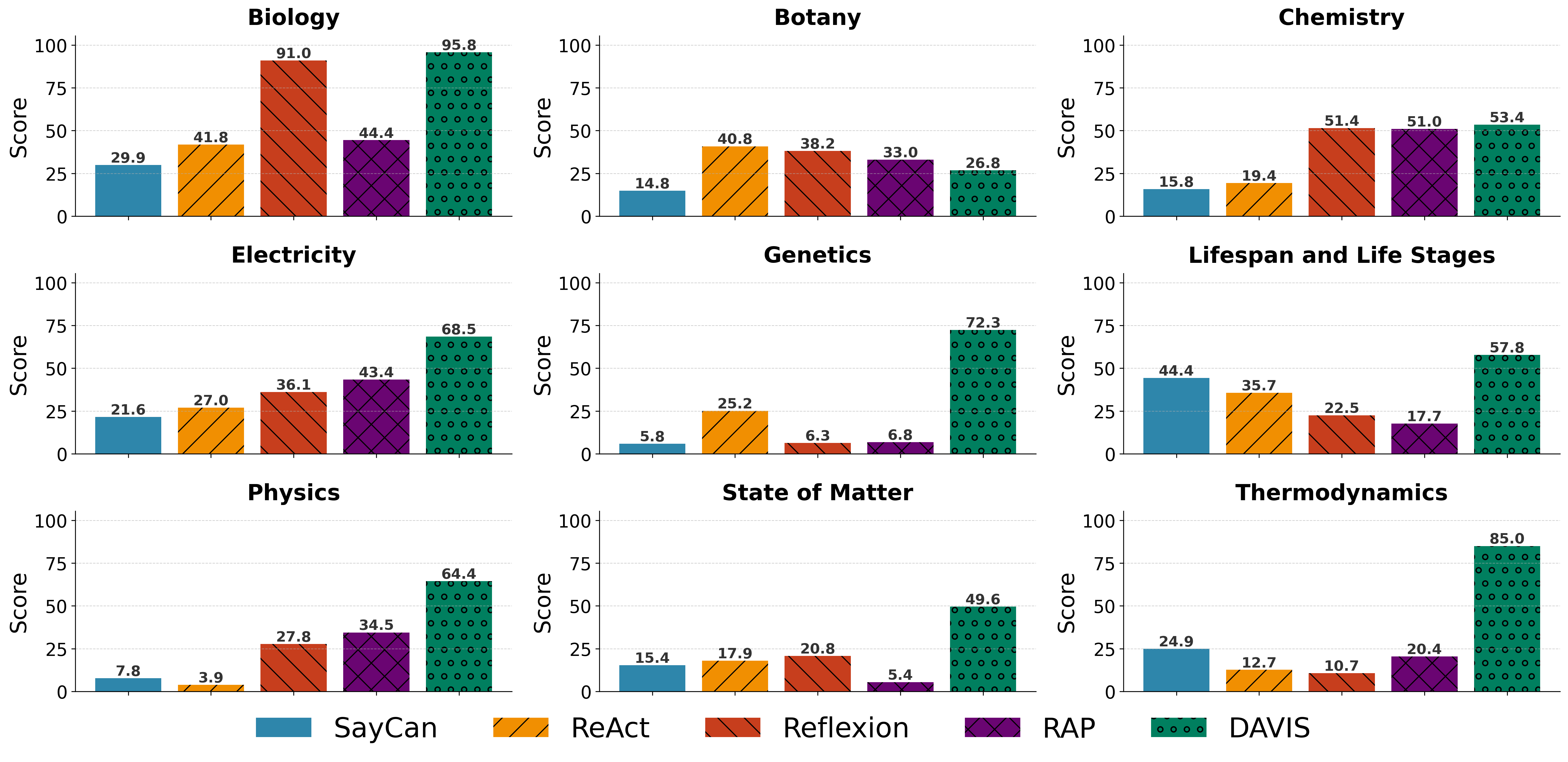}
        \caption{Performance comparison of different agents (SayCan, ReAct, Reflexion, RAP, and DAVIS) across multiple scientific domains. For full results, view table~\ref{tab:full-res} in the Appendix.}
    \label{fig:performance_comparison}
\end{figure*}

The actor-critic system bridges high-level planning and low-level execution. The actor converts the plan into concrete actions in the environment, while the critic evaluates the results and checks whether the outcomes match expectations. If deviations occur, the critic can trigger replanning. This allows for robust operation under partial observability.

\section{Experiment}
\label{sec:evaluation}
\subsection{ScienceWorld environment}
We selected ScienceWorld \cite{wang_scienceworld_2022} as the primary benchmark to evaluate DAVIS, as it is currently the only environment designed for interactive scientific reasoning. It features 30 tasks across 9 grade-school science subjects, set in a simulated lab where agents must navigate 8 functional rooms and use scientific tools to complete tasks. Each task includes more than 100 variations, some of which significantly alter the setup by masking rooms or removing equipment—requiring strong generalization and adaptability. The environment demands common sense reasoning, deduction, and procedural knowledge. Scores reflect progress toward task completion (e.g., 75 indicates 75\% progress before failure), enabling structured and interpretable evaluation. Full details are included in Appendix~\ref{sec:appendix-ScienceWorld}.

\subsection{Performance}
We evaluated DAVIS on the ScienceWorld benchmark, comparing its performance against state-of-the-art baseline agents: SayCan, ReAct, Reflexion, and RAP. The baselines were selected based on their competitive performance, available implementations, and relevance to ScienceWorld. The current state-of-the-art method on ScienceWorld, SwiftSage \cite{lin_swiftsage_2023}, was excluded from our replication baselines because discrepancies between the available code and the documented evaluation methods made direct replication infeasible. For consistency, all baselines were reimplemented to align with the latest ScienceWorld version. For fairness, both RAP and DAVIS utilized memory constructed from five episodes of golden trajectories rather than the ReAct-based approach proposed in \citeposs{kagaya_rap_2024}. Performance was averaged across subjects for comparison, with details on tasks and subjects provided in Tables \ref{tab:tasks} and \ref{tab:classes} in the appendix. Figure~\ref{fig:performance_comparison} shows DAVIS outperforming all baselines in 8 out of 9 subjects, achieving an overall average score of $65.06$—approximately 1.8 times higher than competing methods. Full results for each task, including standard deviations, are in the appendix Table~\ref{tab:full-res}.

Overall, DAVIS took fewer steps before converging to the final score when compared to SayCan, ReAct, and Reflexion. Compared to RAP, DAVIS was better at transferring knowledge from its training to execution despite differences among variations of the same task. Its World Model allows for multi-hop reasoning and inferences based on past training data.


\subsection{Ablation Study}
We systematically evaluate the system with individual modules removed—specifically the World Model (WM), Actor, and Critic—and compare their impact on performance using two metrics: (1) average task score (i.e., task progress before timeout) and (2) average number of steps per replanning cycle (\textit{steps/replan}). To ensure representative coverage across task complexity, we select two tasks from each category of task length (short, medium, and long). For each task, the results are averaged over three different environment variations.

\setlength{\tabcolsep}{1mm}
\begin{table}[t]
\small
\centering 
\begin{tabular}{lcc}
\hline
\textbf{Task} & \textbf{D} & \textbf{D + W} \\
\rowcolor[HTML]{D3D3D3} 
\multicolumn{3}{l}{\textbf{Long Tasks}} \\
\rowcolor[HTML]{D3D3D3} 
Melt (1-2) & 3.00 & 70.00 \\
\rowcolor[HTML]{D3D3D3} 
Determine Melting Point Unk. (2-3) & 5.00 & 92.33 \\
\multicolumn{3}{l}{\textbf{Medium Tasks}} \\
Mix Paint Secondary (6-1) & 40.00 & 36.37 \\
Test Conductivity (3-3) & 55.00 & 58.33 \\
\rowcolor[HTML]{D3D3D3} 
\multicolumn{3}{l}{\textbf{Short Tasks}} \\
\rowcolor[HTML]{D3D3D3} 
Lifespan Longest-Lived (7-1) & 66.67 & 100.00 \\
\rowcolor[HTML]{D3D3D3} 
Find Living Thing (4-1) & 25.00 & 100.00 \\
\hline
\end{tabular}
\caption{DAVIS performance with WM (D + W) and without WM (D) }
\label{tab:WorldModelPerformance}
\end{table}

\paragraph{World Model:}  
Table~\ref{tab:WorldModelPerformance} compares DAVIS with and without its WM. The WM serves as structured external memory in the form of a temporal knowledge graph, enabling grounded and long-horizon planning. Without this, the agent has to rely solely on the internal capabilities of the LLM, lacking access to temporal or multi-hop context.

The WM consistently improves performance across all tasks, particularly in complex and temporally grounded settings like \textit{Melt} and \textit{Find Living Thing}, demonstrating that temporal and structured grounding is critical for high-fidelity decision-making in scientific domains.

\begin{table}[t]
\small
\centering
\caption{Ablation study: Full model (D+W), w/o Actor (D-A), and w/o Critic (D-C). 
}
\label{tab:component-ablation}
\begin{tabular}{llccc}
\toprule
\textbf{Type} & \textbf{Task} & \textbf{D+W} & \textbf{D-A} & \textbf{D-C} \\
\midrule
\multirow{2}{*}{Long} 
& 1-2 & 70 (4.38) & 25 (1.12) & 23.3 (3.20) \\
& 2-3 & 92.3 (1.51) & 79.7 (1.19) & 33.3 (1.29) \\
\midrule
\multirow{2}{*}{Medium} 
& 6-1 & 36.4 (3.22) & 100 (1.49) & 86 (2.49) \\
& 3-3 & 58.3 (2.71) & 28 (1.32) & 49.3 (1.72) \\
\midrule
\multirow{2}{*}{Short} 
& 7-1 & 83.3 (2.00) & 66.7 (2.00) & 83.3 (2.00) \\
& 4-1 & 100 (2.50) & 44.7 (2.94) & 25 (2.50) \\
\bottomrule
\end{tabular}
\end{table}

\paragraph{Actor-Critic.}
We ablate DAVIS in Table~\ref{tab:component-ablation} by removing the Actor or Critic module. In the no-Actor setup, the World Model directly outputs executable actions, skipping high-level goal decomposition. In the no-Critic setup, reflection and subtask updates are disabled, though replanning triggers remain.

\begin{itemize}[left=0pt, itemsep=4pt, parsep=0pt, topsep=4pt]
\item \textbf{No Actor:} The agent struggles to produce valid commands despite access to action formats, resulting in low task scores and near-1.0 steps/replan—indicating constant replanning and poor multi-step coherence. In task 6-1, the agent correctly and luckily guessed the critical action `focus on' within the first 5 steps, skipping all the intermediate steps and resulting in a strong performance in the D-A and D-C agents.

\item \textbf{No Critic:} The agent can execute longer action chains without an Actor module but struggles to recover from errors due to a lack of introspective feedback provided by the Critic. While performance differences are minimal on short tasks—typically solvable within one or two replanning cycles—the gap widens on longer tasks that require more adaptive reasoning. Compared to the no-Actor condition, both task performance and steps per replan improve, but remain below those of the full DAVIS system. 

\end{itemize}

The Actor enables structured execution, and the Critic enhances adaptivity. A higher average steps per replan ratio, paired with strong task scores, demonstrates coherent, cost-efficient planning.

\subsection{Multi-hop Q\&A}
We evaluated the performance of DAVIS’s World Model (WM) on the multi-hop QA benchmarks HotpotQA and MusiqueQA using 400 randomly sampled instances, following the evaluation protocol of \citeposs{li_graphreader_2024}. Table~\ref{tab:wm-benchmark} shows that DAVIS (GPT-4o) achieves strong results, surpassing GraphReader and GraphRAG on HotpotQA with an F1 score of 73.8 and a competitive EM of 56.25, approaching the state-of-the-art HOLMES. On MusiqueQA, DAVIS maintains strong performance (F1: 48.5, EM: 33.8), further demonstrating the effectiveness of its structured, temporal memory in reasoning tasks. While HOLMES achieves the highest overall scores, its static hyper-relational graph architecture lacks DAVIS’s ability to support dynamic updates during inference, which is crucial for agents operating in evolving or interactive environments. Due to costs and the proprietary update schedules of commercial LLMs, we chose to report the best results as reported by the original authors. We note that Table~\ref{tab:wm-benchmark} is not intended to claim state-of-the-art performance but to illustrate how DAVIS’s architectural design generalizes to multi-hop QA tasks. For each system, we report results using the best-performing language model configuration as documented in the respective original papers. We qualitatively observe that retrieval-based systems are highly sensitive to the underlying LLM: DAVIS performs better with GPT-4o than with GPT-4-turbo, despite the latter’s generally stronger performance claims, which should be examined in future work. 



\begin{table}[t]
\small
\centering
\caption{WM comparison against SotA baselines.}
\label{tab:wm-benchmark}
\begin{tabular}{lcc|cc}
\toprule
\textbf{Method} & \multicolumn{2}{c|}{\textbf{HotpotQA}} & \multicolumn{2}{c}{\textbf{MusiqueQA}} \\
 & EM & F1 & EM & F1 \\
\midrule
GPT-4o & 46.3 & 64.1 & 19.0 & 34.4 \\
GPT-4-turbo & 44.3 & 60.4 & 20.5 & 34.7 \\
GraphReader (GPT-4) & 55.0 & 70.0 & 38.0 & 47.4 \\
\textbf{HOLMES (GPT-4)} & \textbf{66.0} & \textbf{78.0} & \textbf{48.0} & \textbf{58.0} \\
GraphRAG (GPT-4o-mini) & 58.7 & 63.3 & 40.0 & 53.5 \\
DAVIS (GPT-4o) & 56.25 & 73.8 & 33.8 & 48.5 \\
DAVIS (GPT-4-turbo) & 55.25 & 71.0 & 34.0 & 47.1 \\
\bottomrule
\end{tabular}
\end{table}

\section{Conclusion}
\label{sec:conclusions}  
DAVIS is an agent designed for scientific interactive reasoning tasks in complex environments. 
DAVIS represents a novel approach that leverages a structured World Model (WM) in the form of a temporal knowledge graph, enabling iterative retrieval and reasoning over past experiences. This structured representation allows DAVIS to approximate both the transition dynamics and reward models of its environment, facilitating more informed decision-making. DAVIS also uniquely uses an interactive retrieval process, which combines iterative querying with contextual reasoning to fill knowledge gaps and refine understanding. This is augmented by DAVIS's ability to perform internal planning and validation before interacting with the environment. By engaging in pre-execution deliberation, DAVIS enables clearer inspection of its planned actions, making it easier for human supervisors to review its decision-making process. This transparency facilitates stronger safeguards compared to reinforcement learning (RL) agents, whose policies are often opaque. DAVIS is ideal for scientific tasks that demand precision, adaptability, and strict adherence to experimental protocols.

Evaluations in the ScienceWorld environment across several scientific domains, including thermodynamics, biology, and physics, demonstrate the efficacy of DAVIS's structured knowledge representation and retrieval methods compared to baseline agents. DAVIS combines robust planning with iterative reasoning capabilities, enabling it to generalize effectively from demonstrations to new tasks. 

\section{Acknowledgments}
The authors acknowledge the support and contributions from members of the Davis Institute for Artificial Intelligence at Colby College in various aspects of the DAVIS project. The authors also extend gratitude to the Accelerate Foundation Models Research initiative for their support. This research was developed with funding from the Defense Advanced Research Projects Agency (DARPA). The views, opinions and/or findings expressed are those of the authors and should not be interpreted as representing the official views or policies of the Department of Defense or the U.S. Government. Assistance with the language of the paper, limited to paraphrasing and polishing authors' original content, was provided by ChatGPT. GitHub CoPilot was used to refine SQL queries, improve LLM prompts, and generate docstrings. Figure~\ref{fig:qa} was created with \href{https://pixelspeechbubble.com/}{https://pixelspeechbubble.com/} and ChatGPT's image generation ability. Figures \ref{fig:main-dia} \& \ref{fig:retrieval-process} were designed in Inkscape. We used graphics from the ScienceWorld paper in Figure~\ref{fig:main-dia}.

\section{Limitations \& Future works}
While DAVIS demonstrates strong reasoning capabilities and improved performance over previous agentic approaches, it has several limitations that should be addressed in future research.

\subsection{High operational cost}
DAVIS relies extensively on Large Language Models (LLMs), resulting in significant computational overhead. Due to its multi-step reasoning process—particularly the structured inner monologue interaction with a Temporal Knowledge Graph; each action involves sending and receiving approximately 43,000 tokens, leading to an average cost of \$0.43 per action. For tasks requiring 100 actions, this can scale to \$43 per episode, totaling over \$3,000 across the 90 experimental variations. While this cost is nontrivial, it reflects a deliberate trade-off between depth of reasoning and operational efficiency. Our goal in this study is to explore the combinations of agent frameworks and language models that best support structured interactive reasoning. To make the experiments more accessible, we use GPT-4o and GPT-4-turbo, which are considerably more affordable than previous GPT-4 models while maintaining stable, high-quality outputs. Looking forward, we see promising opportunities to reduce cost via distillation and fine-tuning of DAVIS's inner monologue architecture onto smaller, open-source models; paving the way for lower-cost, scalable deployments of structured agents in real-world settings.

\subsection{Sensitive to LLM performance}
DAVIS's performance is closely tied to the behavior of the underlying Large Language Models (LLMs), making it sensitive to fluctuations in model quality, API changes, and prompt compatibility. Updates to commercial LLMs can introduce inconsistencies in reasoning accuracy, output structure, and response latency—affecting the agent’s ability to plan and act reliably in dynamic environments. This sensitivity highlights a key limitation of relying on closed-source LLMs for long-term deployment. To address this, future development of DAVIS will prioritize model-agnosticism by integrating smaller, open-source language models.

\subsection{Biased Planning \& Knowledge Dependence}
DAVIS's decision-making process is heavily influenced by the Temporal Knowledge Graph (TKG), which serves as its structured memory. However, this dependence can lead to biased planning, as DAVIS prioritizes information within the graph. Although efforts were made to increase data diversity by populating the knowledge graph with 150 different ScienceWorld task variations, the model still struggles when encountering novel scenarios or incomplete knowledge. Future work should explore adaptive knowledge integration to mitigate bias.

\subsection{Lack of multimodal capabilities, evaluation in safety-critical contexts}
DAVIS operates exclusively in textual environments, limiting its applicability as an embodied agent. The absence of visual, auditory, or sensory perception restricts its ability to interact with real-world multimodal tasks. This is especially necessary in safety-critical contexts. Future research should focus on integrating visual and sensor-based input processing to enhance generalization and deployment in multimodal AI systems. 

Additionally, robust evaluation in real-world and safety-critical contexts is necessary to validate the framework’s reasoning abilities under noisy, high-stakes conditions. This includes benchmarking against embodied QA tasks, robotic planning datasets, and interactive simulations in domains such as remote surgery, chemical lab automation, or aerospace systems. These settings require the agent not only to retrieve and reason, but also to react in real time with awareness of physical constraints, ethical concerns, and safety protocols.

\subsection{Need for Open-Source, Model-Agnostic Evaluation Protocols}

In Table~\ref{tab:wm-benchmark}, we evaluate DAVIS’s World Model (WM) using the HotpotQA and MusiqueQA datasets, reporting F1 and EM scores across systems. To ensure fair comparison, we use results reported by prior work (e.g., GraphReader, HOLMES), rather than re-running those systems. This decision was made due to the high cost of large-scale inference and the instability of commercial APIs, many of which lacks publicly disclosed updates, making exact reproduction infeasible. While informative, this reliance introduces a limitation in terms of reproducibility and transparency.

Moreover, our experiments revealed a high degree of sensitivity to the underlying LLM: DAVIS performed better on QA tasks using GPT-4o than GPT-4-turbo, contrary to general performance claims. Such discrepancies underscore the volatility of relying on commercial LLMs, whose architectures, weights, and behaviors may change without notice.
We believe that this volatility highlights a broader challenge in LLM-based research and advocate for a shift toward open-source, model-agnostic evaluation. Future works should replicate DAVIS’s pipeline using smaller, open-source language models with standardized evaluation protocols, transparent logs, and fixed model checkpoints.

\bibliography{references}

\appendix

\section{ScienceWorld}
\label{sec:appendix-ScienceWorld}
ScienceWorld \cite{wang_scienceworld_2022} is a benchmark designed to evaluate interactive reasoning in digital agents through a realistic laboratory simulation. Developed by the Allen Institute for AI, it provides a text-based environment that emulates scientific experiments, requiring agents to interact with objects, collect observations, and apply reasoning skills to solve tasks. The framework consists of approximately 40,000 lines of \texttt{SCALA} code with a \texttt{PYTHON} interface, following standard RL benchmarking practices.

\begin{figure*}[t]
    \centering
    \includegraphics[width=1\linewidth]{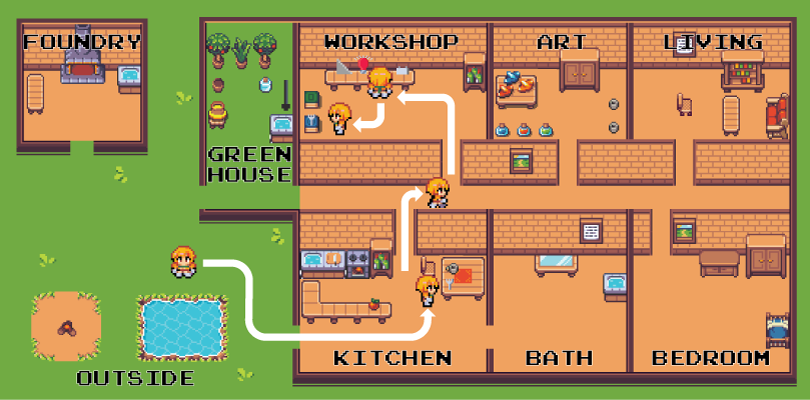}
    \caption{The ScienceWorld environment as presented in \cite{wang_scienceworld_2022}}
    \label{fig:scienceworld}
\end{figure*}

The ScienceWorld environment consists of 10 interconnected locations (Fig.~\ref{fig:scienceworld}), each populated with up to 200 distinct object types, including scientific instruments, electrical components, biological specimens, substances, and common environmental elements such as furniture and books. Agents can interact with objects through a predefined action space of 25 high-level actions, categorized into domain-specific operations (e.g., using a thermometer, measuring conductivity) and general interactions (e.g., moving, opening containers, picking up items). At each step, approximately 200,000 possible action-object combinations exist, though only a subset is relevant based on the context.

ScienceWorld tasks are designed to assess scientific reasoning across multiple disciplines. The dataset includes 30 distinct tasks (Table~\ref{tab:tasks}), covering a range of experimental procedures and problem-solving scenarios. These tasks are further grouped into 9 science domains (Table~\ref{tab:classes}), including physics, chemistry, biology, and environmental science, allowing for targeted evaluation of an agent’s ability to reason through various scientific concepts, making ScienceWorld a robust benchmark for testing multi-step reasoning in dynamic, interactive environments.
\begin{table}[t]
\small
\centering
\vspace{8mm}
\begin{tabular}{ll}
\hline
\textbf{\#} & \textbf{Task} \\ \hline
\rowcolor[HTML]{D3D3D3} 
1-1  & Changes of State (Boiling) \\ 
\rowcolor[HTML]{D3D3D3} 
1-2  & Changes of State (Melting) \\ 
\rowcolor[HTML]{D3D3D3} 
1-3  & Changes of State (Freezing) \\ 
\rowcolor[HTML]{D3D3D3} 
1-4  & Changes of State (Any) \\ 
2-1  & Use Thermometer \\ 
2-2  & Measuring Boiling Point (Known) \\ 
2-3  & Measuring Boiling Point (Unknown) \\ 
\rowcolor[HTML]{D3D3D3} 
3-1  & Create a Circuit \\ 
\rowcolor[HTML]{D3D3D3} 
3-2  & Renewable vs Non-Renewable Energy \\ 
\rowcolor[HTML]{D3D3D3} 
3-3  & Test Conductivity (Known) \\ 
\rowcolor[HTML]{D3D3D3} 
3-4  & Test Conductivity (Unknown) \\ 
4-1  & Find a Living Thing \\ 
4-2  & Find a Non-Living Thing \\ 
4-3  & Find a Plant \\ 
4-4  & Find an Animal \\ 
\rowcolor[HTML]{D3D3D3} 
5-1  & Grow a Plant \\ 
\rowcolor[HTML]{D3D3D3} 
5-2  & Grow a Fruit \\ 
6-1  & Mixing (Generic) \\ 
6-2  & Mixing Paints (Secondary Colours) \\ 
6-3  & Mixing Paints (Tertiary Colours) \\ 
\rowcolor[HTML]{D3D3D3} 
7-1  & Identify Longest-Lived Animal \\ 
\rowcolor[HTML]{D3D3D3} 
7-2  & Identify Shortest-Lived Animal \\ 
\rowcolor[HTML]{D3D3D3} 
7-3  & Identify Longest-Then-Shortest-Lived Animal \\ 
8-1  & Identify Life Stages (Plant) \\ 
8-2  & Identify Life Stages (Animal) \\ 
\rowcolor[HTML]{D3D3D3} 
9-1  & Inclined Planes (Determine Angle) \\ 
\rowcolor[HTML]{D3D3D3} 
9-2  & Friction (Known Surfaces) \\ 
\rowcolor[HTML]{D3D3D3} 
9-3  & Friction (Unknown Surfaces) \\ 
10-1 & Mendelian Genetics (Known Plants) \\ 
10-2 & Mendelian Genetics (Unknown Plants) \\ \hline
\end{tabular}
\caption{\footnotesize Tasks in ScienceWorld.}
\label{tab:tasks}
\end{table}

\section{DAVIS Implementation Details}

We utilized \texttt{GPT-4-turbo} for reasoning, \texttt{GPT-4o} for question answering, and \texttt{LLaMA3-70B} for the Knowledge Graph construction pipeline. Agents were run for a maximum of 80 steps per task. All RAG-based agents were initialized with five variations, a total of 150 variations, of rollouts using the golden trajectory for training, while three randomly sampled test variations, a total of 90 variations, were drawn from the ScienceWorld test set. In contrast, all CoT agents were evaluated directly on the randomly drawn test set as intended.  

All experiments were conducted on a system equipped with a NVIDIA RTX 3060 GPU, an AMD Ryzen 9 7900X CPU, 64GB RAM, running Ubuntu 23.04 with Python 3.11.0. The full table of hyperparameters and settings for DAVIS is provided in Table~\ref{tab:hyperparams}. The results are available in table~\ref{tab:full-res}, and all our code and prompts are available in the attached repository.

\begin{table}[t]
    \centering
    \small
    \renewcommand{\arraystretch}{1.2}
    \begin{tabular}{l|c}
        \hline
        \textbf{Hyperparameter} & \textbf{Value} \\
        \hline
        Maximum Steps per Task & 100 \\
        Simplification Level & Easy \\
        Knowledge Graph Pipeline & LLaMA3-70B-Instruct \\
        Reasoning Model & GPT-4-Turbo \\
        Maximum QA Turns & 5 \\
        Predicted Trajectory Length & 5 \\
        \hline
    \end{tabular}
    \caption{Hyperparameter settings for DAVIS.}
    \label{tab:hyperparams}
\end{table}

\begin{table*}[t]
\begin{tabular}{@{}lll@{}}
\toprule
\textbf{Subject}                                                    & \textbf{Description}                                                                                                                                                  & \textbf{Tasks}                                                   \\ \midrule
\rowcolor[HTML]{D3D3D3} 
Matter                                                              & \begin{tabular}[c]{@{}l@{}}Agents perform experiments to change the state of various\\  materials, such as transforming ice to water or water to steam\end{tabular}   & \begin{tabular}[c]{@{}l@{}}1-1, 1-2,\\ 1-3, 1-4\end{tabular}     \\
Thermodynamics                                                      & \begin{tabular}[c]{@{}l@{}}Agents conduct experiments involving temperature \\ manipulation, such as heating or cooling objects.\end{tabular}                         & \begin{tabular}[c]{@{}l@{}}2-1, 2-2,\\ 2-3\end{tabular}          \\
\rowcolor[HTML]{D3D3D3} 
Electricity                                                         & \begin{tabular}[c]{@{}l@{}}Agents relocate to a workshop and construct electrical\\  circuits to achieve specific objectives.\end{tabular}                            & \begin{tabular}[c]{@{}l@{}}3-1, 3-2, 3-3\\ 3-4\end{tabular}      \\
Biology                                                             & \begin{tabular}[c]{@{}l@{}}Agents relocate to a garden and identify animals\\  based on various queries. \end{tabular} & \begin{tabular}[c]{@{}l@{}}4-1, 4-2, 4-3\\ 4-4\end{tabular}      \\
\rowcolor[HTML]{D3D3D3} 
Botany                                                              & \begin{tabular}[c]{@{}l@{}}Agents relocate to a greenhouse and perform tasks \\ such as growing plants or observing their growth.\end{tabular}                        & 5-1, 5-2                                                         \\
Chemistry                                                           & \begin{tabular}[c]{@{}l@{}}Agents engage in standard chemistry tasks, such as\\ mixing substances to create new compounds\end{tabular}                                & 6-1, 6-2, 6-3                                                    \\
\rowcolor[HTML]{D3D3D3} 
\begin{tabular}[c]{@{}l@{}}Lifespan and \\ Life Stages\end{tabular} & \begin{tabular}[c]{@{}l@{}}Agents observe and report the life stages of plants and\\  animals, such as germination, flowering, or molting.\end{tabular}               & \begin{tabular}[c]{@{}l@{}}7-1, 7-2, 7-3\\ 8-1, 8-2\end{tabular} \\
Physics                                                             & \begin{tabular}[c]{@{}l@{}}Agents use physics knowledge to measure angles or \\ explore physical properties of materials\end{tabular}                                 & \begin{tabular}[c]{@{}l@{}}9-1, 9-2\\ 9-3\end{tabular}           \\
\rowcolor[HTML]{D3D3D3} 
Genetics                                                            & \begin{tabular}[c]{@{}l@{}}Agents identify genetic traits of plants, such as dominant \\ or recessive characteristics, based on observations.\end{tabular}            & \begin{tabular}[c]{@{}l@{}}10-1\\ 10-2\end{tabular}              \\ \bottomrule
\end{tabular}
\caption{\footnotesize Description of subjects and corresponding tasks in ScienceWorld. Each subject represents a unique domain of inquiry, with tasks designed to evaluate agents' reasoning, planning, and execution capabilities in diverse scientific scenarios.}
\label{tab:classes}
\end{table*}

\setlength{\tabcolsep}{2mm}
\begin{table*}[t]
\small
\centering
\begin{tabular}{l|cc|cc|cc|cc|cc}
\toprule
\textbf{Task} & \multicolumn{2}{c|}{\textbf{SayCan}} & \multicolumn{2}{c|}{\textbf{ReAct}} & \multicolumn{2}{c|}{\textbf{Reflexion}} & \multicolumn{2}{c|}{\textbf{RAP}} & \multicolumn{2}{c}{\textbf{DAVIS}} \\
\midrule
 & Mean & Std & Mean & Std & Mean & Std & Mean & Std & Mean & Std \\
\midrule

\rowcolor[HTML]{D3D3D3} \textbf{State of Matter} & \multicolumn{2}{c|}{15.42} & \multicolumn{2}{c|}{17.92} & \multicolumn{2}{c|}{20.83} & \multicolumn{2}{c|}{5.42} & \multicolumn{2}{c}{49.58} \\
1-1 (L) & 1.67 & 1.5 & 2.67 & 2.5 & 27.67 & 41.0 & 13.33 & 20.55 & 25.67 & 19.6 \\
1-2 (L) & 23.33 & 40.4 & 25.67 & 40.2 & 1.00 & 1.7 & 1.67 & 2.89 & 70.00 & 0.0 \\
1-3 (L) & 3.33 & 5.8 & 19.33 & 25.3 & 19.33 & 25.3 & 6.67 & 5.78 & 32.00 & 27.7 \\
1-4 (L) & 33.33 & 57.7 & 24.00 & 39.0 & 35.33 & 56.0 & 0.00 & 0.00 & 70.67 & 0.6 \\

\rowcolor[HTML]{D3D3D3} \textbf{Thermodynamics} & \multicolumn{2}{c|}{24.89} & \multicolumn{2}{c|}{12.67} & \multicolumn{2}{c|}{10.67} & \multicolumn{2}{c|}{20.44} & \multicolumn{2}{c}{85.00} \\
2-1 (M) & 6.00 & 3.0 & 4.00 & 3.5 & 9.00 & 0.0 & 30.33 & 47.43 & 83.00 & 29.4 \\
2-2 (M) & 7.67 & 0.6 & 6.33 & 0.6 & 17.33 & 18.8 & 8.67 & 15.02 & 79.67 & 35.2 \\
2-3 (L) & 61.00 & 48.3 & 27.67 & 39.3 & 5.67 & 0.6 & 22.33 & 20.40 & 92.33 & 13.3 \\

\rowcolor[HTML]{D3D3D3} \textbf{Electricity} & \multicolumn{2}{c|}{21.58} & \multicolumn{2}{c|}{27.00} & \multicolumn{2}{c|}{36.08} & \multicolumn{2}{c|}{43.42} & \multicolumn{2}{c}{68.50} \\
3-1 (S) & 30.33 & 40.4 & 30.33 & 40.4 & 23.33 & 34.5 & 39.00 & 33.05 & 82.33 & 15.7 \\
3-2 (M) & 22.67 & 26.4 & 19.33 & 29.3 & 14.33 & 20.6 & 35.33 & 27.31 & 68.67 & 27.1 \\
3-3 (M) & 23.33 & 27.5 & 5.00 & 5.0 & 39.00 & 34.5 & 38.00 & 35.03 & 58.33 & 2.9 \\

\rowcolor[HTML]{D3D3D3} \textbf{Biology} & \multicolumn{2}{c|}{29.92} & \multicolumn{2}{c|}{41.83} & \multicolumn{2}{c|}{91.00} & \multicolumn{2}{c|}{44.42} & \multicolumn{2}{c}{95.83} \\
4-1 (S) & 11.33 & 9.8 & 17.00 & 0.0 & 72.33 & 47.9 & 61.00 & 38.1 & 100.00 & 0.0 \\
4-2 (S) & 36.00 & 34.8 & 58.33 & 28.9 & 100.00 & 0.0 & 19.33 & 9.8 & 83.33 & 14.4 \\
4-3 (S) & 22.33 & 4.6 & 75.00 & 0.0 & 91.67 & 14.4 & 58.33 & 36.0 & 100.00 & 0.0 \\
4-4 (S) & 50.00 & 43.3 & 17.00 & 0.0 & 100.00 & 0.0 & 39.00 & 38.1 & 100.00 & 0.0 \\

\rowcolor[HTML]{D3D3D3} \textbf{Botany} & \multicolumn{2}{c|}{14.83} & \multicolumn{2}{c|}{40.83} & \multicolumn{2}{c|}{38.17} & \multicolumn{2}{c|}{33.00} & \multicolumn{2}{c}{26.83} \\
5-1 (L) & 16.67 & 14.4 & 9.00 & 3.6 & 3.67 & 4.6 & 50.00 & 73.99 & 35.67 & 2.9 \\
5-2 (L) & 13.00 & 4.6 & 72.67 & 47.3 & 72.67 & 47.3 & 16.00 & 13.89 & 18.00 & 6.2 \\

\rowcolor[HTML]{D3D3D3} \textbf{Chemistry} & \multicolumn{2}{c|}{15.78} & \multicolumn{2}{c|}{19.44} & \multicolumn{2}{c|}{51.44} & \multicolumn{2}{c|}{51.00} & \multicolumn{2}{c}{53.44} \\
6-1 (M) & 16.67 & 11.5 & 23.33 & 11.5 & 56.67 & 37.9 & 53.33 & 5.78 & 36.67 & 5.8 \\
6-2 (S) & 26.33 & 2.3 & 20.67 & 18.0 & 83.33 & 28.9 & 22.67 & 21.60 & 53.67 & 40.5 \\
6-3 (M) & 4.33 & 2.3 & 14.33 & 5.1 & 14.33 & 7.5 & 77.00 & 0.00 & 70.00 & 0.0 \\
\rowcolor[HTML]{D3D3D3} \textbf{Lifespan and Life Stages} & \multicolumn{2}{c|}{44.40} & \multicolumn{2}{c|}{35.67} & \multicolumn{2}{c|}{22.47} & \multicolumn{2}{c|}{17.67} & \multicolumn{2}{c}{57.80} \\
7-1 (S) & 75.00 & 43.3 & 66.67 & 28.9 & 50.00 & 0.0 & 16.67 & 28.86 & 100.00 & 0.0 \\
7-2 (S) & 83.33 & 28.9 & 66.67 & 28.9 & 33.33 & 14.4 & 16.67 & 28.86 & 83.33 & 28.9 \\
7-3 (S) & 33.00 & 0.0 & 22.00 & 19.1 & 22.33 & 9.2 & 5.67 & 9.81 & 83.00 & 0.0 \\
8-1 (S) & 13.33 & 6.1 & 15.00 & 22.6 & 4.00 & 4.0 & 38.00 & 25.98 & 2.67 & 2.3 \\
8-2 (S) & 17.33 & 4.6 & 8.00 & 0.0 & 2.67 & 4.6 & 11.33 & 9.81 & 20.00 & 0.0 \\

\rowcolor[HTML]{D3D3D3} \textbf{Physics} & \multicolumn{2}{c|}{7.78} & \multicolumn{2}{c|}{3.89} & \multicolumn{2}{c|}{27.78} & \multicolumn{2}{c|}{34.48} & \multicolumn{2}{c}{64.44} \\
9-1 (L) & 5.00 & 5.0 & 0.00 & 0.0 & 36.67 & 54.8 & 30.00 & 30.00 & 76.67 & 40.4 \\
9-2 (L) & 6.67 & 7.6 & 11.67 & 12.6 & 8.33 & 2.9 & 30.00 & 0.00 & 60.00 & 34.6 \\
9-3 (L) & 11.67 & 16.1 & 0.00 & 0.0 & 38.33 & 53.5 & 43.44 & 23.28 & 56.67 & 37.9 \\

\rowcolor[HTML]{D3D3D3} \textbf{Genetics} & \multicolumn{2}{c|}{5.83} & \multicolumn{2}{c|}{25.17} & \multicolumn{2}{c|}{6.33} & \multicolumn{2}{c|}{6.83} & \multicolumn{2}{c}{72.33} \\
10-1 (L) & 6.00 & 9.5 & 39.00 & 53.5 & 6.33 & 9.2 & 3.33 & 5.78 & 100.00 & 0.0 \\
10-2 (L) & 5.67 & 9.8 & 11.33 & 9.8 & 6.33 & 9.2 & 10.33 & 10.50 & 44.67 & 47.9 \\

\bottomrule
\end{tabular}
\caption{Full results on ScienceWorld. The average score for each category is displayed in the grey bar on the same row as the category label.}
\label{tab:full-res}
\end{table*}


\end{document}